\documentclass[letterpaper, 10 pt, conference]{ieeeconf}

\IEEEoverridecommandlockouts                              

\usepackage{float}
\usepackage[font=small]{caption}
\DeclareCaptionLabelSeparator{period}{. }
\usepackage{multirow}
\usepackage{epstopdf}
\usepackage{dblfloatfix}
\usepackage{cite}
\usepackage{wrapfig}
\usepackage{listings}
\usepackage{graphicx}
\usepackage{amssymb}
\usepackage{latexsym}
\usepackage{amsfonts}
\usepackage{url}
\usepackage{comment}
\usepackage[linesnumbered,ruled,vlined]{algorithm2e}
\usepackage{algpseudocode}
\usepackage{amsmath}
\usepackage{booktabs}
\usepackage{dcolumn}

{}

\newcommand{\figB}{
\begin{figure}[t]
    \centering
    \includegraphics[width=8cm]{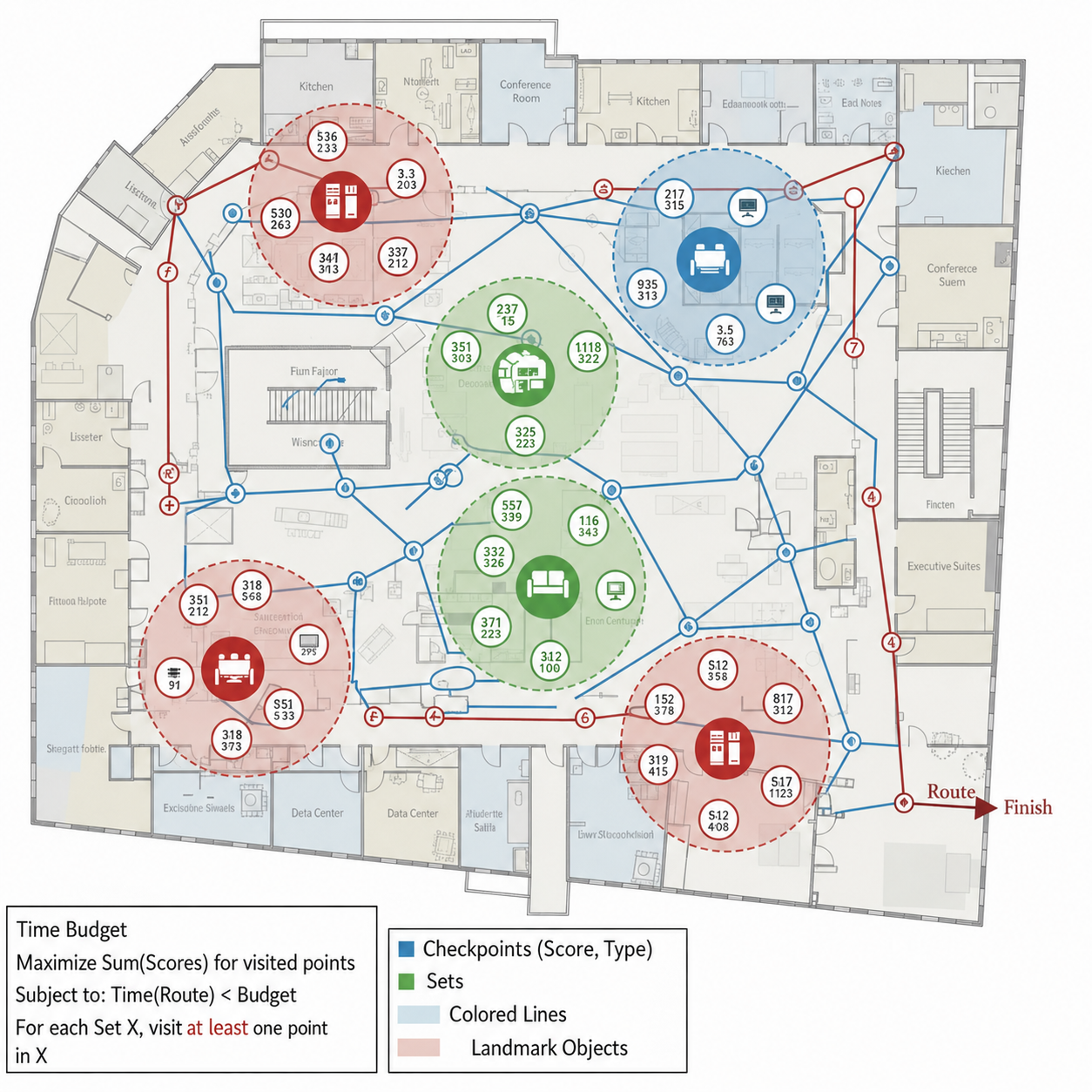}
\vspace{-2mm}
\caption{Problem Formulation: Category-Conditioned Landmark Exploitation as an SOP. Each landmark is enclosed by candidate observation points (colored circles). The agent selects at most one point per set to maximize category search efficiency. Visiting a point provides an observation opportunity; success requires true target detection.}
\vspace{-3mm}
\label{fig:sop_concept}
\end{figure}
}

\newcommand{\figC}{
\begin{figure}[t]
    \centering
    \includegraphics[width=\linewidth]{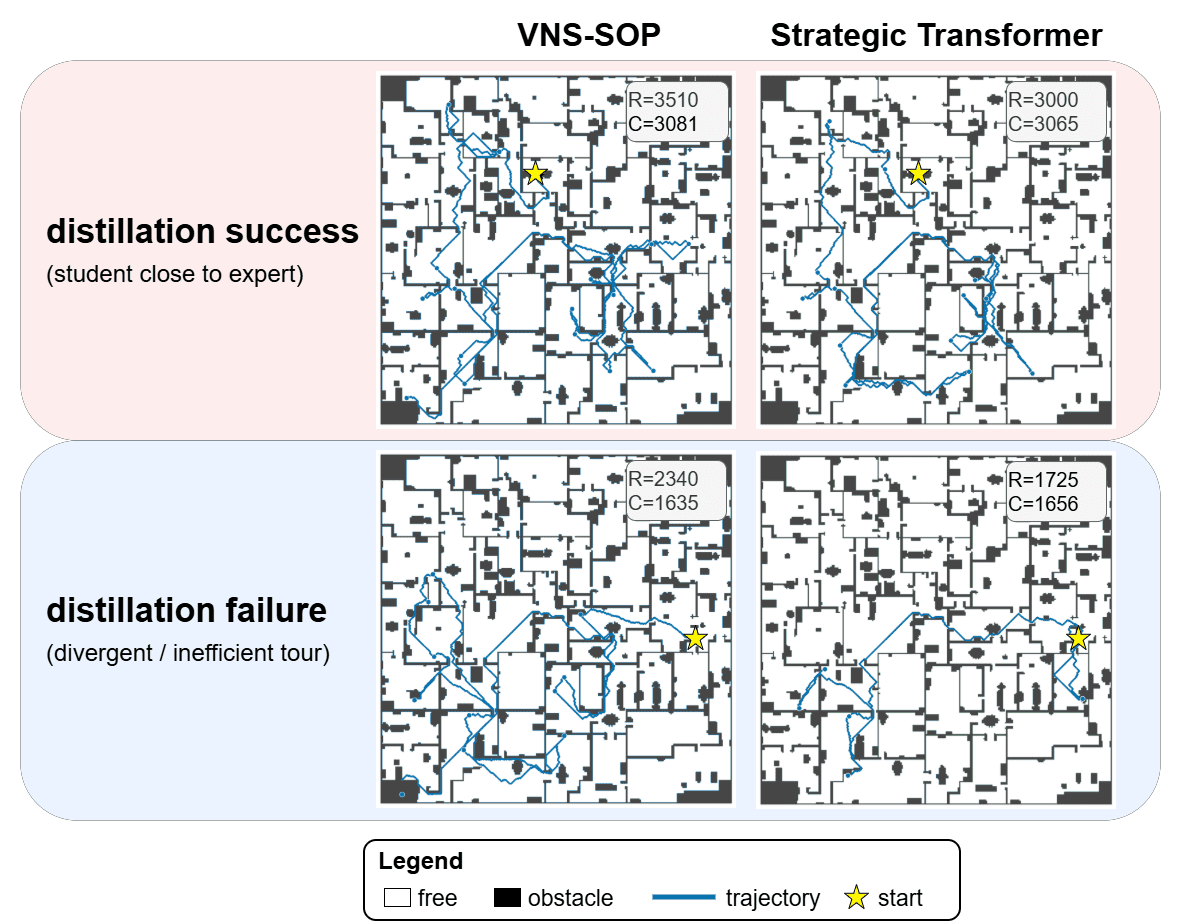}
\vspace{-2mm}
\caption{Qualitative comparison of tours by VNS-SOP (left) and Strategic Transformer + 2-opt (right). Top: distillation success case where the student produces an expert-like tour. Bottom: distillation failure case where the student tour diverges due to overestimating rewards in complex cul-de-sacs. Total reward $R$ and cost $C$ are reported per panel. The failure case yields a 26.3\% lower reward than the expert despite comparable cost.}
\vspace{-3mm}
\label{fig:plan_comparison}
\end{figure}
}

\newcommand{\figD}{
\begin{figure}[t]
    \centering
    \includegraphics[width=\linewidth]{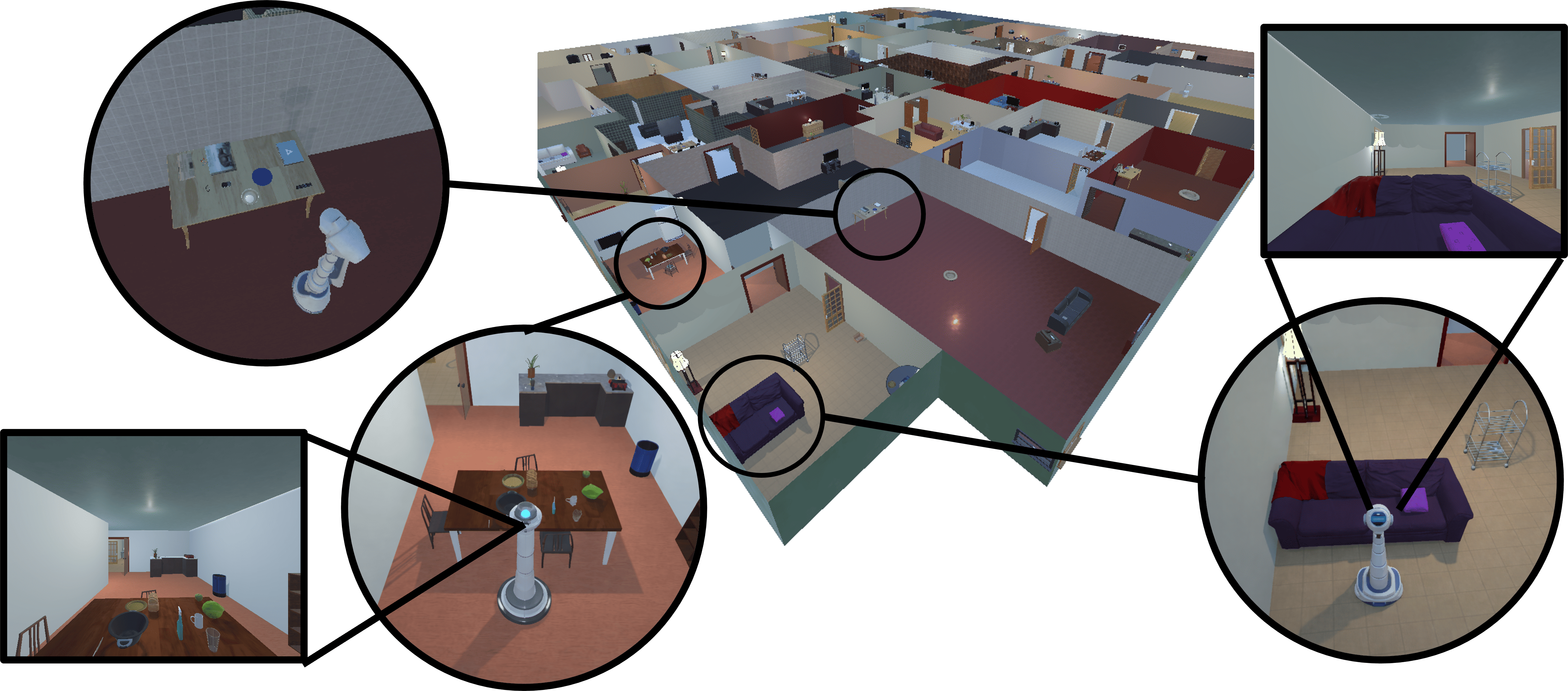}
\vspace{-2mm}
\caption{Examples of target and landmark detections. High-recall visual identification provides the candidate set for strategic SOP planning by clustering semantically relevant viewpoints, even with occasional false positives.}
\vspace{-3mm}
\label{fig:targets_landmarks}
\end{figure}
}

\newcommand{\figE}{
\begin{figure}[t]
\scriptsize
    \centering
    \includegraphics[width=\linewidth]{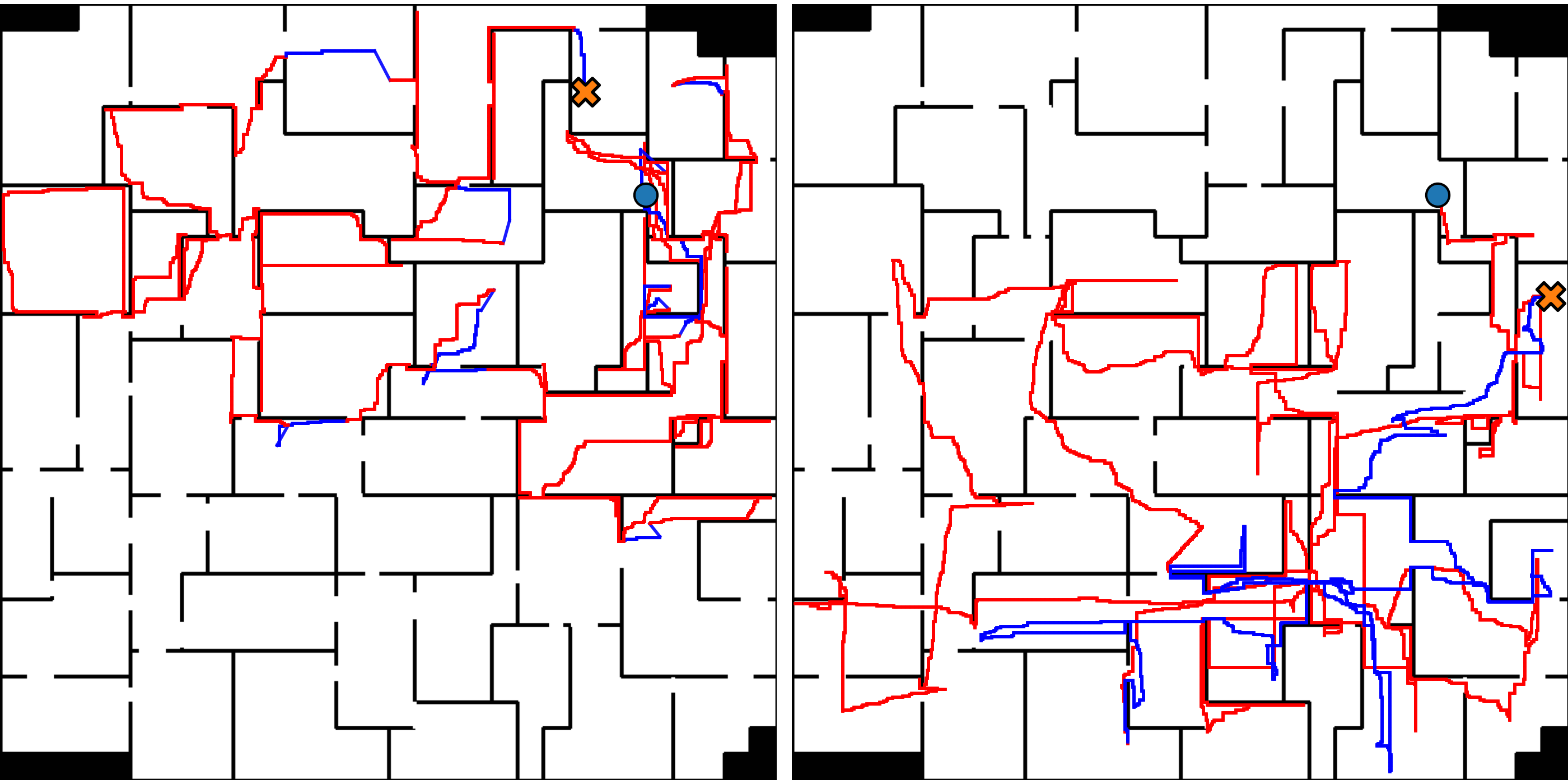}~\\
(a) Conventional method ~~~~~~~~~~~~~~ (b) Strategic Transformer (Ours)
\vspace{-2mm}
\caption{Trajectory comparison in real-scale environments. Red: exploration, blue: touring. Circles/crosses indicate start/end. Conventional methods induce redundant cyclic movements due to frequent mode switching ; the Strategic Transformer suppresses ping-pong behaviors by producing a consistent SOP tour to maximize PPL.}
\vspace{-3mm}
\label{fig:traj_comparison}
\end{figure}
}

\newcommand{\figF}{
\begin{figure}[t]
    \centering
    \includegraphics[width=\linewidth]{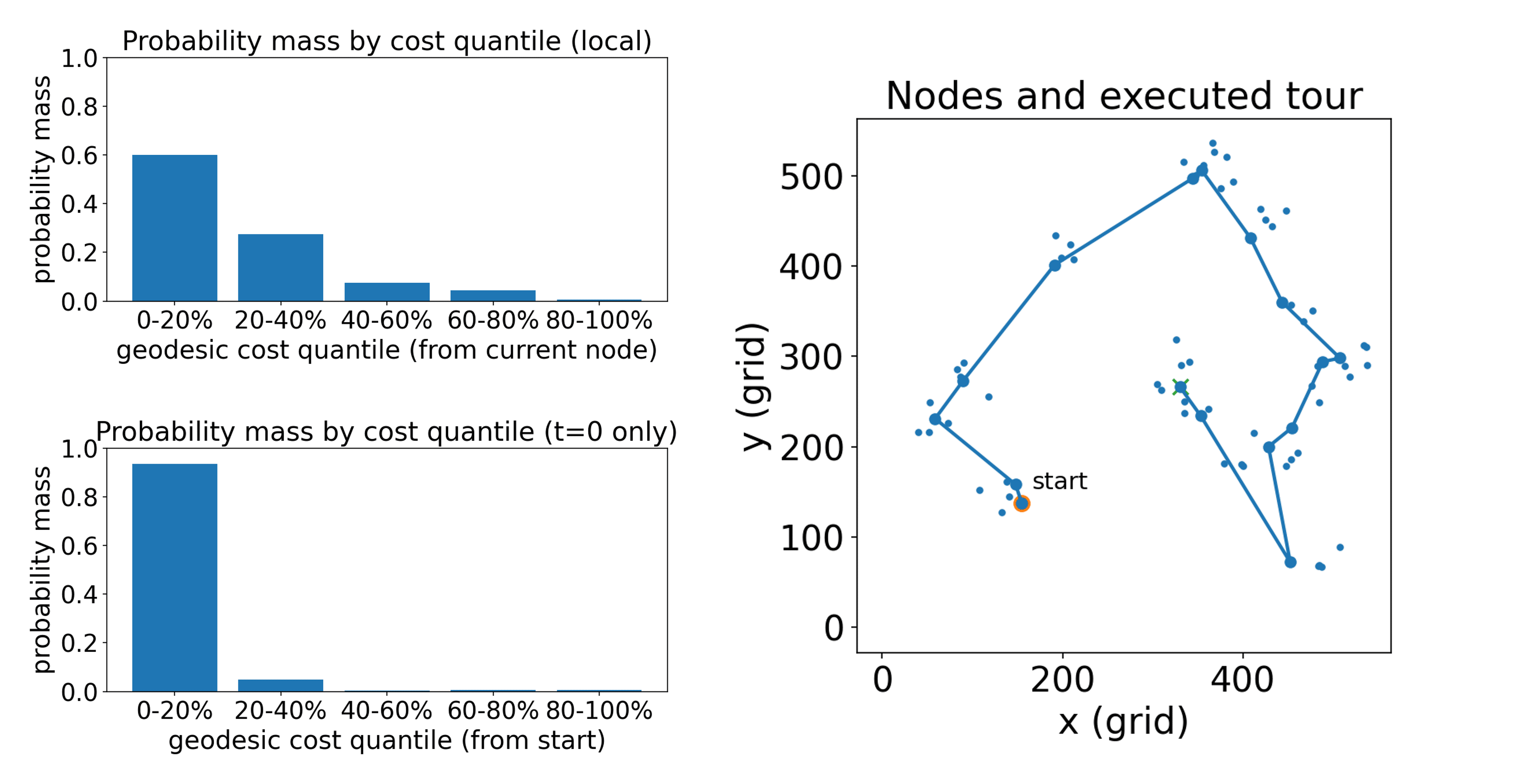}
\vspace{-2mm}
\caption{Distance-dependent probability bias and an example tour. Top left: Candidates are ranked by geodesic cost from the current node and binned into 20\% quantiles ; decoder probability mass is averaged over steps. The distribution is skewed toward low-cost bins, reflecting cost awareness via geometric attention biases. Bottom left: Probability mass at $t=0$ showing a preference for nearby candidates. Right: Executed tour showing a coherent multi-region tour under set constraints.}
\vspace{-3mm}
\label{fig:distance_bias_example}
\end{figure}
}

\begin{document}

\title{\LARGE \bf
Strategic Transformer for Resource-Constrained Multi-Object Navigation in Ultra-Large-Scale Environments
}

\author{Daiki Iwata, Kanji Tanaka, Senta Hishida}

\twocolumn[{%
  \renewcommand\twocolumn[1][]{#1}%
  \maketitle
  \centering
  \includegraphics[width=1.0\textwidth]{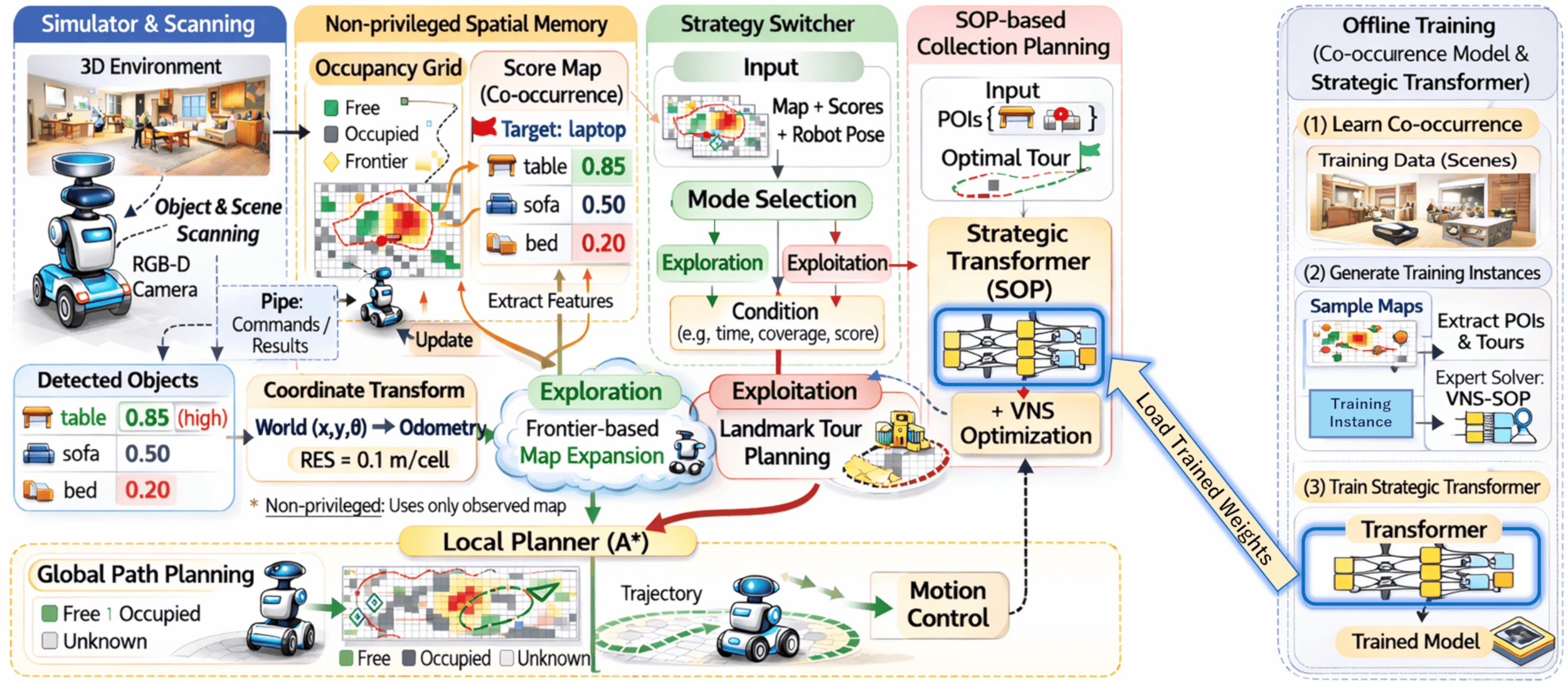}\vspace*{-1mm}\\
  \vspace{-2mm}
  \captionof{figure}{Overview of the BF-SOP architecture, which bridges neural perception and strategic combinatorial optimization. Our framework balances frontier exploration and SOP exploitation via a heuristic strategy switcher, enabling real-time, priority-driven strategic decisions in large-scale, unknown environments via the Strategic Transformer.}
  \vspace{-1mm}
  \label{fig:system_overview}
  \vspace{1.0em} 
}]

\thispagestyle{empty}
\pagestyle{empty}

\makeatletter
\footnotetext[1]{All authors are with the Department of Mechanical Engineering, Faculty of Engineering, University of Fukui, Fukui 910-8507, Japan. (E-mail: tnkknj@u-fukui.ac.jp).}
\makeatother

\begin{abstract}
Resource-constrained multi-object navigation in vast indoor environments ($>2,000\text{ m}^2$) poses significant challenges for efficiency and strategic planning. To tackle this, we reformulate the task as a Set Orienteering Problem (SOP), providing an optimization framework under resource constraints where exploitation is governed by the SOP model and exploration is managed by a separate heuristic switcher. Conventional baselines suffer from either rigid planning or myopic behaviors. To overcome these limitations and resolve the NP-hard computational challenges of SOP for real-time navigation, we develop the Strategic Transformer. This lightweight architecture functions as a priority planner that internalizes expert combinatorial logic into a predictable $41.03\text{ ms}$ forward pass while reducing teacher-student information asymmetry. Incorporating geometric attention biases allows the network to effectively model long-range structural dependencies. By coupling the Transformer's macro-plan with a bounded iterative 2-opt local refinement on a capped candidate graph, our framework achieves a $94\times$ speedup compared to heavy meta-heuristics, ensuring bounded-latency inference suitable for onboard deployment. Experiments on ProcTHOR validate that our method successfully bridges the gap between exploration and exploitation, outperforming carefully re-implemented baselines under Progress weighted by Path Length (PPL) and establishing a new benchmark for scalable, resource-constrained navigation.
\end{abstract}

\section{Introduction}

Resource-constrained multi-object navigation (Multi-ObjectNav) requires an autonomous agent to locate multiple target instances of specific categories within a limited travel resource in unfamiliar environments \cite{Sun2024SurveyObjectGoalNav}. 
While local semantic reasoning and mapping have advanced significantly \cite{Chaplot2020SemExp}, existing research predominantly focuses on small-scale, single-room setups; navigating ultra-large-scale environments ($>2,000\text{ m}^2$) remains a largely unexplored frontier. 
The immense scale of such environments introduces fundamental challenges, as travel trajectories can span multiple kilometers, meaning that a single myopic decision or sub-optimal detour can lead to the irreversible depletion of the entire travel budget. 
In these expansive settings, standard exploration planners often fail due to severe shortsightedness, lacking established global strategies to balance vast spatial exploration with strategic target collection under strict budget constraints. 
This leaves strategic, large-scale navigation challenging, especially since the agent must operate without any prior global map, relying solely on highly restricted online occupancy mapping and real-time pose estimates to make critical macro-planning decisions.

To address these challenges, we propose the \textit{Budget-Frontier-SOP (BF-SOP)} framework, where the planner operates under the same partial information and online mapping constraints as the agent. Our framework decouples operational modes via a heuristic strategy switcher that balances frontier-based exploration \cite{Yamauchi1997Frontier} (intrinsic utility) with the exploitation of detected landmark objects (extrinsic reward). A key core of our approach is formulating candidate revisits during exploitation as a \textit{Set Orienteering Problem (SOP)} \cite{Archetti2018SetOrienteering}, wherein landmarks are defined as SOP cluster center nodes. While a standard Traveling Salesperson Problem (TSP) minimizes cost to visit all nodes, the NP-hard SOP maximizes collected rewards within a travel budget. Critically, we exploit the \textit{set structure} of the SOP to represent multiple candidate viewpoints for a single target object as a cluster, allowing the agent to select an efficient observation point within each set. Fig. \ref{fig:system_overview} illustrates the overall system architecture.

To ensure both logical rigor and computational efficiency during exploitation planning, we introduce the \textit{Strategic Transformer}, which distills the macro-planning logic of an expert solver while reducing teacher-student information asymmetry. Unlike traditional mathematical solvers that suffer from high computational latency in large-scale graphs, our Strategic Transformer internalizes complex sequential prioritization logic into bounded-latency neural inference. By teaching the network the underlying priority-driven planning logic from partial observations, we bridge the gap between global macro-planning and real-time robotic execution.

To establish a comparative baseline for this ultra-large-scale setting, we adapt and evaluate representative ObjectNav frameworks that focus on strategic switching: \textit{BFS-Global} (rigid macro-budget allocation) and \textit{TFL-Reactive} (localized, event-driven reactive approach). However, traditional switching strategies present inherent structural trade-offs: \textit{BFS-Global} completely ignores scene structure, while \textit{TFL-Reactive} suffers from short-sighted triggers that induce a pathological ``ping-pong behavior''—a redundant back-and-forth oscillation that squanders the travel budget.

To overcome the limitations of these conventional strategies, we propose a minimum viable architecture tailored for ultra-large-scale environments ($>2,000\text{ m}^2$), avoiding overly complex heuristics like LLMs to maintain system clarity and efficiency. The primary contributions of this work are threefold: 
1) the \textbf{BF-SOP Planning Framework}, which reduces teacher-student information asymmetry by constraining the teacher to the same online observation and mapping space as the student, balances exploration and exploitation via a heuristic strategy switcher, and manages traversability risk via safety-gated logic; 
2) a \textbf{Conceptual SOP Exploitation Formulation} that models multiple landmark candidates as a conceptual SOP \cite{Penicka2019VNSSetOrienteering}, leveraging the set structure to cluster candidate viewpoints for strategic routing under set constraints; and 
3) the \textbf{Strategic Transformer}, demonstrating that distilling high-latency expert logic under partial-observation constraints into a Transformer-based architecture enables predictable, ultra-low execution latency on a capped candidate graph. By coupling the Transformer's macro-plan with a bounded iterative 2-opt refinement, our framework achieves a $94\times$ speedup, reducing online inference latency to just 41.03 ms. By injecting geometric costs into the attention mechanism \cite{Kool2019AttentionRouting}, this architecture captures structural dependencies to achieve substantial computational acceleration with a manageable quality gap. We evaluate our method under Progress weighted by Path Length (PPL) in the ProcTHOR framework \cite{Deitke2022ProcTHOR}, where it significantly outperforms our re-implemented switching baselines.
Our code and baseline re-implementations will be made available on GitHub upon acceptance.

\section{Related Work}

\subsection{The Scalability Gap in Resource-Constrained Multi-ObjectNav}
While autonomous exploration originated with Frontier-Based Exploration (FBE) \cite{Yamauchi1997Frontier}, the challenges of \textit{ultra-large-scale} resource-constrained category-conditioned object search remain an unaddressed frontier \cite{Sun2024SurveyObjectGoalNav}. Current deep reinforcement learning and LLM-based approaches \cite{Chaplot2020SemExp, Petroni2019LMasKB} exhibit myopic behavior in environments exceeding $2,000\text{ m}^2$, prioritizing local rewards over global strategic planning \cite{Mirowski2017LearningToNavigate}. Unlike Success weighted by Path Length (SPL), which is traditionally used in single-object setups and only measures binary success, Progress weighted by Path Length (PPL) \cite{wani2020multion} is uniquely suited for multi-target scenarios as it reflects partial success and discovery efficiency under strict resource constraints.

Our approach redefines mode activation by analyzing existing strategic-switching frameworks, specifically focusing on \textit{BFS-Global} (rigid macro-budget allocation lacking adaptability to scene layouts) and \textit{TFL-Reactive} (localized, event-driven reactive approach inducing pathological oscillations). To establish a fair comparison in this ultra-large-scale setting, we re-implement these baselines within a unified ProcTHOR framework. Our proposed strategy switcher operates on a global macro-level horizon, evaluating budget states and expected long-term returns to suppress unstable state-toggling, allowing the Strategic Transformer to execute far-sighted strategic coordination with a predictable, ultra-low online inference latency profile.

\subsection{Reducing Information Asymmetry via Partial-Observation Learning}
Early imitation learning utilized ``privileged oracles'' with full environmental knowledge for supervision \cite{cite_dagger}, a trend continued by planners using ground-truth maps to accelerate convergence \cite{Wijmans2020DDPPO}. However, this creates a critical information asymmetry since such priors are unavailable during real-world inference \cite{symmetry2024}. Our work bridges this gap through partial-observation learning that reduces teacher-student information asymmetry. The expert builds a virtual environment based on the full map constructed retrospectively after the agent finishes exploring. By simulating multiple ObjectNav tasks within this virtual environment, the expert solver collects diverse decision-making trajectories offline under partial observations, which are then used for distilling the Strategic Transformer, establishing a logically consistent learning environment.

\subsection{Distilling Combinatorial Optimization for Bounded-Latency Planning}
To the best of our knowledge, we are the first to formulate and validate \textit{ultra-large-scale} ObjectNav as a resource-constrained optimization problem via the Set Orienteering Problem (SOP), bridging the gap between embodied AI and Operations Research \cite{Tsiligirides1984Orienteering, Archetti2018SetOrienteering}. 
Although Graph Neural Networks (GNNs) are commonly utilized to approximate NP-hard problem solutions \cite{Joshi2019GCN_TSP}, they completely fail in this task because the underlying graph topology of the unexplored environment is not known \textit{a priori}, preventing effective spatial message passing. 
To address this, our Strategic Transformer requires no predefined topology and leverages global self-attention to model sequential prioritization trade-offs between distant candidates. 
The heavy mathematical computations of the VNS solver are confined strictly to offline training. 
During online execution, the Strategic Transformer completely replaces the solver to internalize expert logic into a predictable, low-latency priority planner, ensuring an efficient computational footprint suitable for onboard deployment.

\section{Approach}

\subsection{Graph Construction from Online Observations}
The agent incrementally builds a semantic graph $G_{obs} = (V, E)$ from past trajectories and egocentric observations, assuming no prior global map \cite{elfes1989occupancy}. The maps and candidates are updated at each decision step $t$.

\subsubsection{Node Representation and Dual-Mode Graph}
$G_{obs}$ expands dynamically with $V$ partitioned into two distinct operational modes:
(i) \textbf{Exploitation Mode ($V_{\text{target}} \cup V_{\text{viewpoint}}$):} Targets target object categories via sets of candidate viewpoints $S_i$ generated on the partial map. In our conceptual SOP formulation \cite{Archetti2018SetOrienteering}, each candidate set is associated with a detected landmark object $L_i$ (an SOP cluster center) which serves as a semantic cue for exploration rather than the target itself. Here, the semantic co-occurrence probability used to identify these landmarks is a statistical correlation of object arrangements calculated based on the diverse pre-simulation trajectory data accumulated by the offline expert, ensuring it does not rely on privileged external knowledge. Crucially, visiting $v \in S_i$ merely provides an observation opportunity for physical verification rather than guaranteeing a successful target find; actual discovery is strictly validated only upon true target detection by the agent's sensors. To ensure tractability, we retain only top landmarks of interest (LOIs). 
(ii) \textbf{Exploration Mode ($V_F$):} Targets frontiers between explored and unexplored space \cite{Yamauchi1997Frontier} to provide intrinsic utility for map expansion and entropy reduction.

This dual-mode architecture avoids unstable reactive mode-toggling. Instead, a heuristic strategy switcher operates as a top-level decision algorithm using the remaining budget, visit costs, and exploration rates. It evaluates whether the current state warrants a transition from map expansion to structured multi-target exploitation. By accounting for the global remaining budget and long-term predicted returns simultaneously, the strategy switcher suppresses unstable behavior. Once scheduled, the Strategic Transformer's long-horizon plan preserves global route consistency, bounding mode activation by resource-aware optimization rather than local geometric fluctuations.

\subsection{Conceptual Variable-Horizon SOP Formulation}
Strategic decision-making within the exploitation mode is defined as a non-privileged variable-horizon problem over $G_{obs}$. Unlike local reactive policies, this seeks a global macro-touring plan for landmark exploitation under partial observability (Fig. \ref{fig:sop_concept}).

\figB

\subsubsection{Conceptual SOP Objective and Constraints}
To model the exploitation of multiple candidate sets, we introduce a conceptual SOP objective aimed at prioritizing candidate nodes under a given travel cost convention \cite{Tsiligirides1984Orienteering}:
\begin{equation}
\max_{x,z,y} \ w_R \sum_{S_k\in \mathcal{S}} r_k y_k - w_C \sum_{i,j\in V} c_{ij}x_{ij}
\end{equation}
subject to $\sum_{i,j\in V} c_{ij}x_{ij} \le B$, $\sum_{j\in V}x_{ij} = \sum_{j\in V}x_{ji} = z_i$ ($\forall i\in V$), $\sum_{i\in S_k}z_i = y_k$ ($\forall S_k\in \mathcal{S}$), and $x_{ij},z_i,y_k\in\{0,1\}$, where $c_{ij}$ is the edge cost, and $z_i, y_k$ are binary indicators. We exploit the \textit{set structure} \cite{Carrabs2021BRKGAforSOP}, where $S_k$ is a cluster of viewpoints; $y_k=1$ if any $i \in S_k$ is visited. This allows efficient entry point selection and forms the basis for PPL, with $L_{\mathrm{opt}}$ computed as the minimum cost to visit discovered targets. This conceptual SOP model is used exclusively for routing within the exploitation mode; although it mathematically incorporates frontier exploration and intrinsic rewards within its structural formulation, the operational activation and transition between modes are managed and isolated by the external heuristic strategy switcher to maintain architectural clarity and roles separation.

We solve this NP-hard problem using Variable Neighborhood Search (VNS) \cite{Penicka2019VNSSetOrienteering} as our expert solver. While VNS yields high-quality solutions, its heavy computational latency makes it unsuitable for real-time execution. Crucially, the VNS solver is restricted to offline training data generation and is completely deactivated during online navigation, where our lightweight Transformer network predicts solutions to eliminate online optimization overhead.

\subsection{Strategic Transformer Architecture}
\label{subsec:strategic_transformer}

To enable frequent replanning on the graph $G_{obs}$ without online expert overhead, we introduce the \textit{Strategic Transformer}, a lightweight neural priority planner. While preliminary experiments showed that Graph Neural Networks (GNNs) produce myopic, locally-optimal paths due to neighborhood-based aggregation, the Transformer architecture leverages global self-attention to evaluate long-horizon trade-offs between distant high-reward candidates. Under our minimum viable architecture, the network functions as a budget-independent priority planner, while budget constraints are enforced via external truncation gates and bounded iterative local refinement on a capped candidate graph. This substitution delivers significant computational acceleration, ensuring a predictable, low-latency profile well-suited for real-time deployment.

A key feature is the \textbf{Biased Attention} mechanism. Instead of relying solely on coordinate embeddings, we inject the normalized geodesic cost matrix $C$ (calculated via A*) directly into the self-attention layers as a geometric bias \cite{ying2021graphormer}. This conditions the attention computation on the traversable cost structure of the navigation graph, providing geometric context beyond coordinate embeddings. Rather than assigning end-to-end continuous constraints to an oversized neural network, our minimalist design satisfies them through deterministic logic rules. The Strategic Transformer captures the macro-combinatorial logic, while immediate structural constraints are governed by a bounded iterative 2-opt alignment and a predictable safety gate, managing computational complexity and ensuring clear performance attribution.

To implement this complete system, the Strategic Transformer serves as the foundational macro-planner integrated with a bounded iterative 2-opt refinement on a capped candidate graph. Our 2-opt execution evaluates node pair exchanges up to a fixed maximum iteration count, bounding the worst-case refinement latency relative to the capped graph size. This module performs both \textit{inter-set sequence reordering} and \textit{intra-set node exchange} (dynamically swapping a viewpoint $v_i \in S_k$ with an alternative candidate within the same cluster $S_k$), optimizing local path efficiency while maintaining bounded-latency decoding consistency.

\subsubsection{Spatial-Semantic Context Embedding}
Each node in the graph is represented by a feature vector $\mathbf{x}_i = [\tilde{p}_{i,x},\ \tilde{p}_{i,y},\ \tilde{r}_i]$, where $(\tilde{p}_{i,x},\tilde{p}_{i,y})$ are the normalized 2D coordinates within the current online map and $\tilde{r}_i \in [0,1]$ is the normalized reward inherited from the corresponding landmark set. This feature vector is linearly projected to a latent embedding and processed by the Transformer encoder.

\subsubsection{Geometric Attention Bias}
To incorporate the geometric structure of the navigation graph into representation learning, the encoder takes a pairwise travel-cost matrix $C=[c_{ij}]$ computed via A* search on the online occupancy map \cite{Hart1968AStar}. This cost matrix is mapped to an attention bias term and directly added to the scaled dot-product attention matrix, explicitly penalizing high geodesic distances and enforcing spatial awareness during the self-attention calculation.

This cost matrix $C$ is injected as an \emph{additive bias} to the self-attention scores. To guarantee bounded-latency execution regardless of scale, the number of inputted landmark candidates is strictly capped. Under partially observed environments, we adopt an \textit{optimistic planning strategy} where unobserved cells are temporarily treated as traversable space to calculate $c_{ij}$. This ensures that matrix generation and subsequent 2-opt refinement operate within predictable latency limits, preventing computational bottlenecks.

This mechanism enables all-to-all node interactions while explicitly incorporating the instance-specific geodesic cost structure via $b_{ij} = \log(\hat{c}_{ij}+\epsilon_{dist})$, where $\hat{c}_{ij} = c_{ij} / \max(m,\epsilon_{dist})$, $m = \max_{u,v} c_{uv}$, and $\epsilon_{dist}=10^{-9}$. Crucially, $b_{ij}$ conditions the encoded representations so that information aggregation reflects geodesic distance variations on the semantic map, allowing the model to prioritize strategically essential nodes even if they are far in Euclidean space.

\subsubsection{Policy Distillation Head}
On top of the encoder, an autoregressive pointer-style policy head \cite{Vinyals2015PointerNetworks} produces a distribution over candidate vertices. The head computes dot-product scores against all encoded node embeddings and applies a masking strategy to prevent revisiting vertices and ensure set diversity (selecting at most one viewpoint per landmark set). During runtime, greedy decoding selects the next vertex. 

This Transformer-based policy head acts as a high-level priority planner. Training utilizes behavioral cloning under teacher forcing \cite{cite_dagger} to internalize the expert's sequential prioritization logic into bounded-latency neural inference.

\subsubsection{Inference, Computational Complexity, and Execution}
During online navigation, an SOP instance is constructed by selecting the top $N_{\text{LOI}}=15$ LOIs and generating $N_{\text{vp}}=5$ viewpoints per LOI. The value of $N_{\text{LOI}}=15$ is experimentally determined as the maximum threshold where the offline SOP expert solver can solve the instance accurately within an acceptable operational time for training data generation.

While the VNS-based solver iteratively executes local search with 2-opt thousands of times within its optimization loop, making online execution prohibitive, the Strategic Transformer generates an optimized skeleton of the tour via a single-shot forward pass in just a few milliseconds. 

Following this macro-planning, a bounded iterative SOP-adapted 2-opt refinement is applied once as a lightweight post-processing step on the capped candidate graph. This module optimizes local path efficiency by performing both inter-set sequence reordering and intra-set node exchange to reduce the total travel cost. Because the graph and iteration counts are strictly capped, the entire computation—including A* distance matrix generation—faithfully guarantees predictable real-time capability and avoids combinatorial tail-latency variance \cite{Archetti2018SetOrienteering}.

During online deployment, physical travel limits and traversability risks are managed via distinct gating mechanisms. A \textbf{Budget Gate} monitors real-time odometry and terminates tour execution when the accumulated travel cost reaches $B$. Separately, a \textbf{Traversability Gate} detects navigation risks in unobserved or complex regions based on the online grid map.

\section{Experiments}

\subsection{Experimental Setup}

\subsubsection{Environment and Task}
\label{subsec:task_definition}
We evaluate the framework in the \textbf{AI2-THOR} simulator \cite{Kolve2017AI2THOR} using \textbf{ProcTHOR} \cite{Deitke2022ProcTHOR}, focusing on ultra-large-scale residential environments ($>2,000\text{ m}^2$) with approximately 50 rooms. The task is resource-constrained Multi-ObjectNav, where an episode is evaluated based on a single target category that may contain multiple distinct instances scattered throughout the environment. Formally, the agent aims to maximize the unique discovery progress within a total travel budget resource constraint $B$:
\begin{equation}
\max_{\pi} \ |\mathcal{F}_{T}| \quad \text{subject to} \quad \sum_{t=1}^{T} c(a_t) \le B
\label{eq:task_objective}
\end{equation}
where $\mathcal{F}_{T} \subseteq \mathcal{O}^{\star}$ denotes the set of unique target instances found up to time $T$, $\mathcal{O}^{\star}$ represents all instances of the assigned target category, and $c(a_t)$ is the travel cost of action $a_t$.

The evaluation protocol encompasses 90 validation episodes across 9 distinct target object categories listed in Table \ref{tab:target_categories} under predefined travel budget limits. To isolate planning performance, the navigation stack utilizes simulator-provided pose data and instance-segmentation masks.

\begin{table}[h]
\centering
\caption{Target Object Categories Evaluated in the Experiments}
\label{tab:target_categories}
\setlength{\tabcolsep}{4pt} 
\begin{tabular}{cll}
\hline
ID & Category Name & Operational Subtype \\ \hline
1 & Apple & Small Item / Countertop \\
2 & Bread & Small Item / Countertop \\
3 & Cup & Small Item / Multipurpose \\
4 & Laptop & Electronics / Desk \\
5 & Book & Medium Item / Shelf \\
6 & Potato & Small Item / Kitchen \\
7 & SoapBottle & Small Item / Bathroom \\
8 & Tomato & Small Item / Kitchen \\
9 & HousePlant & Decorative / Floor-Furniture \\ \hline
\end{tabular}
\end{table}

\subsubsection{Baselines and Re-implementation}
To establish a rigorous benchmark in the ProcTHOR environment, we re-implemented two representative switching paradigms under a standardized sensor configuration for fair comparison. Performance is evaluated using Progress weighted by Path Length (\textbf{PPL}) \cite{wani2020multion}:
\begin{enumerate}
    \item \textbf{BFS-Global}: A baseline utilizing a fixed, rigid macro-budget threshold to transition from exploration to exploitation, completely independent of local scene structure or environmental complexity.
    \item \textbf{TFL-Reactive}: A localized, event-driven reactive baseline that toggles modes based on myopic cues such as elapsed time or close landmark proximity, frequently inducing resource-wasting ping-pong behaviors.
\end{enumerate}

\subsubsection{Observation and Mapping}
Visual observations are processed via a tiered semantic parsing pipeline with an egocentric $90^{\circ}$ Field of View (FoV). To suppress noise, any detected instance with a spatial footprint under 100 pixels is filtered out:
\begin{equation}
\text{Ignore region if } A_c < 100\text{ pixels}
\label{eq:noise_filter}
\end{equation}
where $A_c$ denotes the pixel area of the detected target region. Valid objects exceeding 400 pixels are categorized as primary landmarks of interest (LOIs) for strategic SOP planning, provided they statistically co-occur with the target category:
\begin{equation}
\text{Recognize as Landmark if } A_c \ge 400\text{ pixels}
\label{eq:landmark_recognition}
\end{equation}
A target instance is registered as discovered in $\mathcal{F}_t$ if it enters the FoV within a physical distance of 1.0 m. The proximity threshold $\theta_c$ for initiating verified interaction with target category $c$ is defined as:
\begin{equation}
\theta_c = 1000 s_c
\label{eq:proximity_threshold}
\end{equation}
where $s_c$ is the base spatial scale factor of category $c$.

The agent maintains a top-down metric 2D occupancy grid and a traversability map with a resolution of $\Delta = 0.1\text{ m/pixel}$. The occupancy grid is updated dynamically via ray casting. For each ray $r$ projected from the sensor origin $p_{\text{agent}}$, the log-odds occupancy value $L(m_i)$ of a grid cell $m_i$ is updated as:
\begin{equation}
L(m_i \mid z_t) = L(m_i \mid z_{t-1}) + \text{inverse}(z_t)
\label{eq:raycasting_update}
\end{equation}
where $\text{inverse}(z_t)$ returns a positive log-odds value for occupied cells and a negative value for free space cells. To keep the planning graph tractable and bound decoding latency, we retain the capped $N_{\text{LOI}}=15$ LOIs and generate $N_{\text{vp}}=5$ reachable viewpoints around each candidate using an $A^*$ algorithm.

\subsection{Neural Component Configurations}
The Strategic Transformer consists of an encoder-decoder architecture with 6 encoder layers, 8 attention heads, and a latent embedding dimension of $d=128$ \cite{Vaswani2017Attention}. The network is trained via behavioral cloning under teacher forcing \cite{cite_dagger} using the Adam optimizer with a learning rate of $1 \times 10^{-4}$ and a gradient clipping threshold of $1.0$, employing a step-based learning-rate decay schedule over 100 epochs. 

Training labels are generated using a Variable Neighborhood Search (VNS) algorithm \cite{Penicka2019VNSSetOrienteering} as the non-privileged expert. To generate this large-scale expert dataset, the VNS-SOP optimization problem is solved offline over procedural ProcTHOR layouts to construct a dataset of approximately $10^5$ step-level state--action transitions. For the training pipeline and baseline benchmarks, the expert logic is supported by a high-performance external SOP solver compiled in C++, which interfaces with the simulation controller via high-throughput file-based pipe communication.

To ensure consistent runtime execution, the pairwise matrix generation and subsequent 2-opt refinement are strictly constrained to a maximum of 100 iterations on the capped graph. All training and evaluations were conducted on an NVIDIA RTX 3090 GPU.

\subsection{Baselines}
To evaluate the proposed BF-SOP framework, we compare it against baseline switching heuristics adapted for the AI2-THOR/ProcTHOR environment \cite{Kolve2017AI2THOR, Deitke2022ProcTHOR}:
1) \textbf{Nearest Frontier (FBE)} \cite{Yamauchi1997Frontier}: A classical geometry-driven map expansion strategy navigating to the nearest frontier, which inherently lacks any capability to exploit semantic target information.
2) \textbf{BFS-Global}: A baseline utilizing a fixed macro-budget threshold to transition from exploration to exploitation, independently of local map modifications.
3) \textbf{TFL-Reactive}: An event-driven reactive baseline that toggles behavioral modes based on immediate temporal and spatial landmark proximity triggers.
4) \textbf{BF-SOP (Ours - VNS Teacher)}: Our framework configured with an offline VNS solver \cite{Penicka2019VNSSetOrienteering} for explicit SOP optimization, serving as an upper bound (deactivated during online deployment due to heavy latency constraints).
5) \textbf{BF-SOP (Ours - Strategic Transformer)}: Our fast priority planner distilling VNS expert logic into a Strategic Transformer \cite{Kool2019AttentionRouting}.

\subsection{Evaluation Metrics}
To assess strategic planning performance under resource constraints, we adopt the Progress weighted by Path Length (PPL) metric \cite{wani2020multion}. Rather than employing a binary success definition typical of single-object navigation tasks, our evaluation measures the unique instance discovery progress over a multi-instance landscape. Specifically, the base progress metric $P_T$ at the end of an episode $T$ is defined as the ratio of unique discovered target instances to the total number of existing instances of the assigned category in the environment:
\begin{equation}
P_T = \frac{|\mathcal{F}_T|}{|\mathcal{O}^{\star}|}
\label{eq:unique_progress_metric}
\end{equation}
where $\mathcal{F}_T$ is the set of unique discovered target instances and $\mathcal{O}^{\star}$ represents the total ground-truth instances of that specific target category present in the environment. The final PPL score is computed by weighting this unique discovery progress $P_T$ by the path length efficiency factor $\min(1, L_{\text{opt}} / L)$, where $L$ is the actual path length executed by the agent, and $L_{\text{opt}}$ is the optimal geodesic path distance required to retrospectively traverse and visit only the subset of instances that were successfully discovered during the episode.

\subsubsection{Execution and Qualitative Assessment}
During online evaluation in \textbf{AI2-THOR}, the inference latency of the Strategic Transformer is measured on an NVIDIA RTX 3090 GPU to verify real-time execution constraints. Physical budget limits and navigational risks are handled dynamically by the gating layers as detailed in Section III. A qualitative comparison of the generated routes, including an analysis of distillation failure cases where the model overestimates rewards in complex cul-de-sacs, is illustrated in Fig. \ref{fig:plan_comparison}.

\figC

\begin{table}[t]
\vspace{-2mm}
\caption{Comparison of the average PPL score in ultra-large-scale environments}
\vspace{-1mm}
\label{tab:overall_scores_avg}
\scriptsize
\begin{center}
\begin{tabular}{lc}
\hline
Method & Average PPL $\uparrow$ \\
\hline
Nearest Frontier (FBE) \cite{Yamauchi1997Frontier} & 0.1059 \\
BFS-Global (Rigorously Re-implemented) & 0.1796 \\
TFL-Reactive (Rigorously Re-implemented) & 0.1862 \\
\textbf{BF-SOP (VNS Teacher - Offline Expert Only)} & \textbf{0.3231} \\
\textbf{BF-SOP (Strategic Transformer)} & \textbf{0.2258} \\
\hline
\end{tabular}
\end{center}
\vspace{-3mm}
\end{table}

\begin{table}[t]
\vspace{-2mm}
\caption{Standalone performance of the extrinsic planner (SOP solver) and solver latency}
\vspace{-1mm}
\label{tab:sop_solver_compare}
\begin{center}
\scriptsize
\setlength{\tabcolsep}{2pt} 
\begin{tabular}{@{} l r r r r @{}} 
\hline
Method & Avg Reward & Avg Cost & Time (ms) & Speedup \\
\hline
Transformer-SOP + 2-opt & 2014.5 & 2279.7 & \textbf{41.03} & \textbf{94.11$\times$} \\
VNS-SOP \textbf{(Offline Training Only)} & 2753.2 & 2216.8 & \textbf{3861.42} & 1$\times$ \\
\hline
\end{tabular}
\end{center}
\vspace{-3mm}
\end{table}

\subsection{Quantitative Results}

\subsubsection{Navigation Performance in Ultra-Large-Scale Environments}
Evaluating on the ultra-large-scale ProcTHOR benchmark shows that the proposed BF-SOP framework outperforms state-of-the-art switching heuristics by a clear margin (Table~\ref{tab:overall_scores_avg}). While the offline VNS Teacher establishes the performance upper bound, BF-SOP with the Strategic Transformer successfully internalizes the core macro-sequential prioritization logic. 

Crucially, the evaluation verifies that our method efficiently bridges the gap between exploration and long-horizon multi-target exploitation, whereas traditional paradigms fail under these unprecedented scales. The Strategic Transformer simultaneously overcomes the macro-blindness of \textbf{BFS-Global} and suppresses the myopic volatility of \textbf{TFL-Reactive}, achieving a balanced coordination that drastically improves PPL efficiency.

\subsubsection{Efficiency and 2-opt Ablation}
The core planning latency of our framework shows substantial efficiency gains (Table~\ref{tab:sop_solver_compare}). The Strategic Transformer achieves an approximate $94\times$ speedup over the high-latency VNS-SOP expert solver, reducing the online execution latency to 41.03 ms. This demonstrates that far-sighted strategic decisions can be executed with the same computational lightness as short-sighted reactive heuristics.

¼Furthermore, integrating the bounded iterative 2-opt local refinement on the capped candidate graph successfully yields a 6.4\% travel cost reduction (minimizing the average travel cost from 2435.6 to 2279.7) while increasing computational overhead by a negligible 0.16 ms (Table~\ref{tab:two_opt_ablation}). This confirms that restricting local refinement to a capped graph strips away heavy iterative complexity while preserving local optimization benefits for real-time control loops, providing a predictable execution profile without combinatorial tail-latency variance.

\begin{table}[t]
\centering
\vspace{-2mm}
\caption{Effectiveness of 2-opt post-processing}
\vspace{-1mm}
\scriptsize
\label{tab:two_opt_ablation}
\setlength{\tabcolsep}{6pt}
\begin{tabular}{lcc}
\toprule
Method & Avg. Cost $\downarrow$ & Time (ms) \\
\midrule
Transformer-SOP (w/o 2-opt) & 2435.6 & \textbf{40.87} \\
Transformer-SOP (w/ 2-opt)  & \textbf{2279.7} & 41.03 \\
\bottomrule
\end{tabular}
\vspace{-3mm}
\end{table}

\figF
\figD
\figE

\subsection{Qualitative Analysis and Strategic Consistency}

\subsubsection{Distance-Conditioned Probability Bias}
To examine how the Strategic Transformer reflects geometric travel costs, we visualize the decoder's next-node selection probabilities aggregated by distance quantiles. In Fig.~\ref{fig:distance_bias_example} (Top left), probability mass concentrates in near-cost bins (0--20\% and 20--40\%) and decreases for farther bins, demonstrating that the Geometric Attention Bias successfully conditions representation learning based on geodesic distance variations. Fig.~\ref{fig:distance_bias_example} (Bottom left) shows a preference for nearby candidates at the first step ($t=0$). Crucially, this bias does not induce purely greedy behavior; as shown in the qualitative example (Fig.~\ref{fig:distance_bias_example}, Right), the policy produces a coherent multi-region tour that satisfies set constraints and optimizes sequences. This indicates the model internalizes the expert's sequential prioritization logic, balancing global coherence with locally efficient transitions. Target categories and landmarks are visualized in Fig. \ref{fig:targets_landmarks}. Simulator-provided observations provide the basis for identifying landmarks, and despite occasional false positives in complex layouts, high recall ensures goal-relevant objects are integrated into the SOP plan.

\subsubsection{Trajectory Comparison in Real-Scale Environments}
Through the trajectory analysis shown in Fig.~\ref{fig:traj_comparison}, the structural vulnerabilities of conventional switching strategies become evident. \textit{BFS-Global} fails to adapt to environment modifications, while \textit{TFL-Reactive} triggers excessive mode-toggling that induces resource-wasting ping-pong behaviors. In contrast, our Strategic Transformer suppresses these redundant oscillations by producing a globally consistent tour, preventing budget waste and ensuring high exploration efficiency.

\subsubsection{Trajectory Comparison in Real-Scale Environments}
Fig.~\ref{fig:traj_comparison} compares executed trajectories between baseline switching heuristics and our method without a prior global map. Through this analysis, the structural vulnerabilities of traditional paradigms become evident. Due to its fixed allocation approach, \textbf{BFS-Global} is entirely blind to localized scene modifications, failing to adjust resources dynamically. Conversely, \textbf{TFL-Reactive} suffers from severe short-sightedness; lacking an integrated global plan, its mode-toggling induces a pathological ``ping-pong behavior.'' As captured in Fig.~\ref{fig:traj_comparison}(a), the agent erratically backtracks through previously traversed regions, squandering its travel budget.

In stark contrast, the Strategic Transformer (Fig.~\ref{fig:traj_comparison}(b)) completely suppresses these near-sighted oscillations. By simultaneously evaluating the remaining budget and long-term expected returns, our priority planner assigns a far-sighted touring order over candidate nodes, forming a coherent movement pattern resilient to local geometric fluctuations. This ensures that frontier expansion and target collection progress efficiently, avoiding redundant back-and-forth movements and driving the higher PPL reported in our quantitative evaluation.

\section{Concluding Remarks}
\label{sec:conclusions}

This work presented the \textbf{BF-SOP} framework, bridging global strategic macro-planning and execution in ultra-large-scale resource-constrained multi-object navigation. By reformulating landmark exploitation as a conceptual Set Orienteering Problem \cite{Archetti2018SetOrienteering} and leveraging the \textbf{Strategic Transformer} with geometric attention biases \cite{ying2021graphormer}, we achieved bounded-latency decision-making on a capped candidate graph. 

Our framework successfully overcomes both the macro-blindness of traditional global thresholds and the myopic volatility of reactive event-driven strategies. By substituting heavy metaheuristics with our lightweight neural priority planner, the framework eliminates online optimization overhead while delivering far-sighted strategic coordination suitable for real-world onboard deployment. Experiments on ProcTHOR \cite{Deitke2022ProcTHOR} validated that our macro-planning framework significantly outperforms baseline switching heuristics under the PPL metric, achieving substantial computational speedups with a manageable quality gap. Future work will extend the Strategic Transformer to dynamic environments and integrate multi-modal reasoning for nuanced landmark selection under sensory noise.

\bibliographystyle{IEEEtran}
\bibliography{references}

\end{document}